\documentclass[letterpaper, 10 pt, conference]{ieeeconf}  
\IEEEoverridecommandlockouts                              
\usepackage{graphicx}
\usepackage{subfigure}
\usepackage{algorithm}
\usepackage[noend]{algpseudocode}
\usepackage{amsmath}
\usepackage[space]{cite}
\usepackage{chngpage}

\title{\LARGE \bf
A Stereo Algorithm for Thin Obstacles and Reflective Objects
}

\author{ John Keller* and Sebastian Scherer*
\thanks{* All  authors  are  with  the  Robotics  Institute  at  Carnegie Mellon  University,  Pittsburgh,  Pennsylvania,  USA.
{\tt\small [jkeller2, basti]@andrew.cmu.edu}}
}

\begin{document}
\maketitle
\thispagestyle{empty}
\pagestyle{empty}
 
\begin{abstract}

Stereo cameras are a popular choice for obstacle avoidance for outdoor lighweight, low-cost robotics applications. However, they are unable to sense thin and reflective objects well. Currently, many algorithms are tuned to perform well on indoor scenes like the Middlebury dataset. When navigating outdoors, reflective objects, like windows and glass, and thin obstacles, like wires, are not well handled by most stereo disparity algorithms. Reflections, repeating patterns and objects parallel to the cameras' baseline causes mismatches between image pairs which leads to bad disparity estimates. Thin obstacles are difficult for many sliding window based disparity methods to detect because they do not take up large portions of the pixels in the sliding window. We use a trinocular camera setup and micropolarizer camera capable of detecting reflective objects to overcome these issues. We present a hierarchical disparity algorithm that reduces noise, separately identify wires using semantic object triangulation in three images, and use information about the polarization of light to estimate the disparity of reflective objects. We evaluate our approach on outdoor data that we collected. Our method contained an average of 9.27\% of bad pixels compared to a typical stereo algorithm's 18.4 \% of bad pixels in scenes containing reflective objects. Our trinocular and semantic wire disparity methods detected 53\% of wire pixels, whereas a typical two camera stereo algorithm detected 5\%.

\end{abstract}

\section{INTRODUCTION}

Stereo cameras are frequently used for 3D perception in robots navigating outdoor environments. In order to be robust for this application, stereo depth estimation algorithms need to be real time and handle challenging scenes that contain thin objects, repeating patterns, objects parallel to the cameras' baseline, and reflective surfaces. Currently, many stereo depth estimation algorithms are tuned to perform well in artificial indoor scenes, such as the popular Middlebury dataset.

To address these challenges, we developed a stereo algorithm that is optimized to handle repeating patterns using a hierarchical method to provide more context without a significant increase in running time. To handle objects parallel to the baseline of the stereo pair, we use a trinocular setup with the third camera added perpendicular to the baseline. This creates two perpendicular camera baselines, so ambiguities caused by objects parallel to one camera pair's baseline will be resolved by the perpendicular camera pair. To address reflective surfaces, one of the cameras in the trinocular setup is a micropolarizer camera, which is capable of simultaneously capturing an image at four polarization angles. This information allows reflective surfaces, like water and glass, to be identified and surface normals to be estimated. In scenes with reflective objects, our method contained 9.27\% of bad disparity pixels whereas a typical stereo algorithm contained 18.4\%.

We handle thin wire objects using a trinocular algorithm and semantic wire triangulation. We identify wires using semantic segmentation and estimate their disparity using images collected from a trinocular camera setup. Our method detected 53\% of wire pixels, whereas a typical two camera stereo algorithm detected 5\%.

Our contribution on handling reflective objects is to create a real time algorithm that uses a micropolarizer camera to identify and estimate the depth of specular image regions. Since the of depth specular regions is found by fitting the regions into the disparity image, it can be used in conjuction with any dense disparity algorithm. Previous work was not real time and was designed to work with one disparity algorithm. Our contribution on estimating the depth of thin obstacles is to use semantic segmentation to limit the search area for matching features in images, which simplifies the matching problem. Previous works use depth estimates to refine image based semantic segmentation, whereas we use image based semantic segmentation to refine depth estimates.

Section \ref{rw} discusses related work. Sections \ref{stereo}, \ref{trinocular}, and \ref{polarcam} describe our stereo, trinocular, and methods for handling reflective surfaces. Results and conclusions are presented in Section \ref{results} and Section \ref{conclusion}.

\begin{figure}[h]
	\centering
	\includegraphics[width=0.9\linewidth]{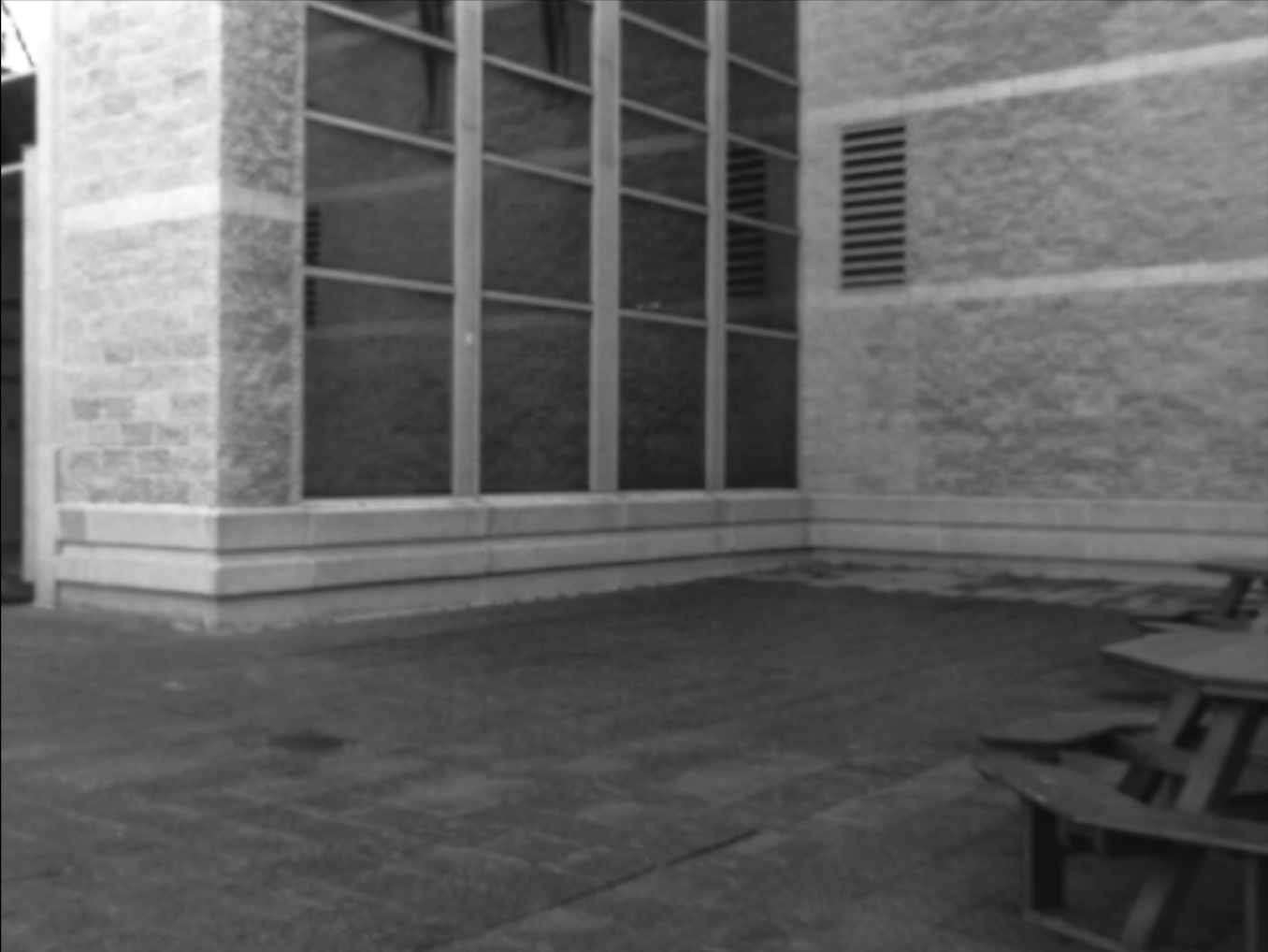}
	\caption{A challenging scene for typical disparity algorithms featuring a reflective window.}
	\label{fig:scene}
\end{figure}

\section{RELATED WORK} \label{rw}

Estimating depth using stereo cameras is done by finding the disparity between corresponding pixels in multiple images taken from cameras at known positions. Dense or sparse disparity can be computed. In both cases, images are rectified so that pixels corresponding to the same object are in the same row in both images. Dense disparity is found at each pixel in an image by examining a neighborhood window around a pixel in the base image and finding the window in the match image that is most similar. There are numerous cost functions for measuring the similarity between windows of pixels. The most commonly used cost function is the sum of absolute differences (SAD) where the absolute difference of corresponding pixels in a window are summed. Another common cost function is to find the absolute difference between interpolated pixel values \cite{cost2}. Both of these methods are sensitive to global radiometric differences between images, since they are directly comparing pixel intensities. To overcome this, the Census transform cost function compares the relationship between pixel intensities with the center pixel of a window, rather than comparing the raw intensities \cite{cost1, cost3}. The Census transform forms a bit string with 1s or 0s based on whether each pixel in the window has a greater or lesser intensity than the center pixel. The hamming distance between bit strings at pixels in both images is computed to measure similarity. Since there are global differences between pixel intensities in the micropolarizer and regular grayscale cameras, the Census cost function is ideal for handling this.

Given scores for a range of disparities at each pixel, the minimum cost disparity can be chosen to form a disparity image. This often leads to a noisy image, with large variations between disparity of adjacent pixels. To refine the disparity, a smoothness penalty is added that penalizes fast changes in disparity, and some optimization is done to minimize the penalty over the image. This can be done globally, as is done in Graph Cuts \cite{gc}, however, this is computationally expensive and is not real time. For real time scenarios, semi-global matching (SGM) is used \cite{sgm}. SGM iterates across horizontal, vertical, and diagonal paths in an image and uses dynamic programming to minimize the disparity cost and smoothness penalty in within each path.

Less noisy disparity estimates can also be obtained by incorporating more context into the window similarity measurement. This can be done by increasing the size of the window or by using the same window size on a hierarchy of image sizes. Hierarchical methods compute disparity scores at low image resolutions first and use the best scoring disparities to bound the disparity search range at higher resolutions \cite{hier1} \cite{hier2}. This improves computational time by limiting the disparities searched through and incorporates more of the image into the cost of each pixel. One disadvantage is that the correct disparity is not within the bounds set at a lower resolution, it will not be found. Another disadvantage is that small objects will not contribute much to the disparity cost at lower resolutions, which increases the likelihood they will not be detected. One way to address this is by downsampling the image in a way that preserves fine details \cite{hier_thin}. To overcome these disadvantages, instead of bounding the disparity ranges searched for at higher resolutions using lower resolutions, we interpolate the lower resolution costs up and add them to the higher resolution costs. We also handle thin objects separately.

Alternatively, sparse disparity can be computed. Instead of finding the disparity at every pixel, disparity is only found at feature points in the image. Feature points can be identified using one of many feature detectors, like SIFT, SURF, or ORB, and matched using feature descriptors. Instead of points, edge features can also be used \cite{thin}. A disadvantage of using edges as features is that in high texture regions, which are common in natural outdoor environments, there can be a large amount of edges with a similar appearance which makes matching difficult. To overcome this we use semantic wire detection \cite{wire} to identify regions of the image containing wires, and only match edge pixels from these regions. Previous semantic segmentation methods have incorporated sparse \cite{seg_sparse} and dense \cite{seg_dense} depth estimates. Since the depth to thin objects is difficult to estimate, instead of using depth estimates as additional information in semantic segmentation, we use semantic segmentation to provide additional information in estimating depth. This allows us to get the benefits of smooth disparity estimates from the dense disparity, while detecting thin objects with semantic wire detection.

One common issue with two camera stereo configurations is that objects which are parallel to the baseline of the cameras lead to inaccurate disparity estimation. This is because the costs of disparities along the edge of the object will be almost identical to each other, so image noise will tend to be the deciding factor on which disparity costs the least. To address this another camera can be added to the configuration that is perpendicular to the baseline of the original two cameras. This ensures that objects cannot be parallel the baselines of both camera pairs. To incorporate all three camera images into the cost function, one camera image is chosen as the base image. The matching costs between each of the other two cameras are then summed \cite{tri1}. This can also help disambiguate repeating patterns. The costs from each pair of images can also be weighted at each pixel based on which image pair is more likely to give reliable costs \cite{tri3}. In real world scenarios wires are often horizontal, so adding a third camera perpendicular to a horizontal camera pair leads to more accurate estimation of the disparity of wires.

Another common source of error in disparity images are reflections caused by specular surfaces. If the reflections are clear as in a glass window or water surface, this can lead to the disparity being estimated to the object in the reflection rather than the surface causing the reflection. In \cite{reflect} an estimate of surface reflectivity is built into the disparity cost function and is used to reduce the noise caused by reflections but not entirely eliminate it and is not real time. Alternatively, a PolarCam can be used to detect areas of the image with specular reflection. The PolarCam has a polarization filter over the image sensor, which polarizes incoming light at each pixel at either 0, 45, 90, or 135 degrees. This gives information about how light is polarized which can be used to estimate the degree of linear polarization (DOLP) and estimate the surface normal of the object. In \cite{polar}, the surface normal estimates from this camera were used on specular objects combined with dense disparity estimates in non specular regions, to estimate the disparity of the image using a global method. It is not real time, but leads to accurate disparity in specular regions. We propose an alternative method that is real time.

\section{STEREO} \label{stereo}

Our stereo disparity algorithm is designed to produce smooth disparity images and reduce false positives caused by repeating patterns and objects parallel to the baseline. To do this we use a hierarchical approach, three image resolutions are used, full size, half size, and quarter size. Pseudocode for our algorithm is given in Algorithm 1. We chose to use the Census cost function because there are some global pixel intensity differences between the micropolarizer camera and regular graysgraycale cameras, even when the exposures are chosen to be the same. We also apply SGM on the sum of hierarchical disparity costs. Instead of using the hierarchical costs to constrain the disparity ranges of higher resolution images, we interpolate the scores from the lower resolution images and sum them together. This means that information from lower resolution is still incorporated, but if the disparity is incorrect at lower resolutions it will not force the higher resolutions to take on these costs. Since more context around a pixel is being considered, this can lead to depth discontinuities at object borders to be expanded and some thin object being missed. Since our application is obstacle avoidance for outdoor robots object borders being expanded is not a major issue since many obstacle avoidance algorithms artificially expand obstacles in order to plan around them. In order to detect thin obstacles the window size could be reduced, however, this increases noise since less image context is considered. Thin objects, such as wires or branches, are difficult to perceive in window based algorithms because the background tends to take up more pixels in the window than the thin objects. Since modifications to window based algorithms to detect thin obstacles usually involve less image context being considered, we instead have a separate step that identifies thin objects using edge detection and estimates their depth with trinocular cameras. We also use a common way of filtering out potentially inaccurate disparity values. If the lowest cost disparity is within a certain percentage of the second lowest cost non adjacent disparity then the cost is considered to not be unique, so the disparity at the pixel is invalidated.

\begin{algorithm}
\caption{Calculate a cost for each disparity of an image using a hierarchy of images.}
\hspace*{\algorithmicindent}\textbf{Input:} $I_l$ The left image.

\hspace*{\algorithmicindent}
$I_r$ The right image.

\hspace*{\algorithmicindent}
$d_{max}$ The maximum disparity range to search.

\hspace*{\algorithmicindent}
$GetDisparityCosts$ A function taking the left and right images and maximum disparity which returns a cost for each row, column and disparity of the image.

\begin{algorithmic}

\State $Costs[max\_scale] = []$

\For{$i \gets 1, max\_scale$}

\State $Costs[i] =$
\State $GetDisparityCosts[resize(I_l, i), resize(I_r, i), d_{max}]$

\EndFor

\For{$i \gets max\_scale-1,...,1$}
\For{$r \gets 0, rows/i$}
\For{$c \gets 0, cols/i$}
\For{$d \gets 0, max\_disparity/i$}

\State $cost_1 = Costs[i-1, r/i, c/i, d/i]$
\State $cost_2 = Costs[i-1, r/i, c/i, d/i+1]$
\State $Costs[i, r, c, d] += \dfrac{cost_1 + cost_2}{2}$

\EndFor
\EndFor
\EndFor
\EndFor

\Return Costs[1]
\end{algorithmic}
\end{algorithm}

\section{TRINOCULAR} \label{trinocular}

The same cost function, hierarchical method, and SGM are used for the trinocular cameras. Rather than summing the costs from both pairs of images together, we use the method discussed in \cite{tri3} to weight the costs from each image pair. This weight is designed to favor the cost from the image pair whose baseline is perpendicular to objects.

\subsection{Semantic Wire Triangulation}

In order to find the disparity of wires, we first use a dilated convolutional neural network trained to detect wires \cite{wire}. It takes as input a color image and outputs a probability for each pixel indicating how likely it is to belong to a wire. The probabilities are thresholded which identifies small regions around wires. In order to extract the pixels belonging to the edge of the wire so that they can be matched between images we find the overlap between the wire regions and edges detected using a Canny edge detector. This is done for all three images, then edge pixels are matched between images using the following cost function for the right, top image pair:
$$ Cost_v(x, y, d) = $$
$$(|I_r(x, y) - I_t(x, y+d)| + |V_r(x, y) - V_t(x, y+d)|)*(V_r(x, y)) $$

and the left, right image pair:
$$ Cost_h(x, y, d) = $$
$$(|I_r(x, y) - I_l(x+d, y)| + |H_r(x, y) - H_l(x+d, y)|)*(H_r(x, y)) $$

where $I$ is the image, $V$ and $H$ are the vertical and horizontal gradients, and the subscripts $r$, $l$, and $t$ indicate the right, left, and top images. This takes the sum of the absolute difference of pixel intensities and the absolute difference of gradients. The vertical stereo pair cost is weighted by the vertical gradient, and the horizontal stereo pair cost is weighted by the horizontal gradient. This means that costs will be penalized more if the gradient of the edge is closer to being parallel with the camera pair's baseline. SGM is then applied by iterating along the edges starting at their endpoints. The same uniqueness criteria mentioned above is also applied to the edge disparity costs. The lowest cost disparity value is chosen at each edge pixel and added to the trinocular disparity image.

\section{REFLECTIVE SURFACES} \label{polarcam}

\begin{figure}[b]
	\centering
	\includegraphics[width=0.9\linewidth]{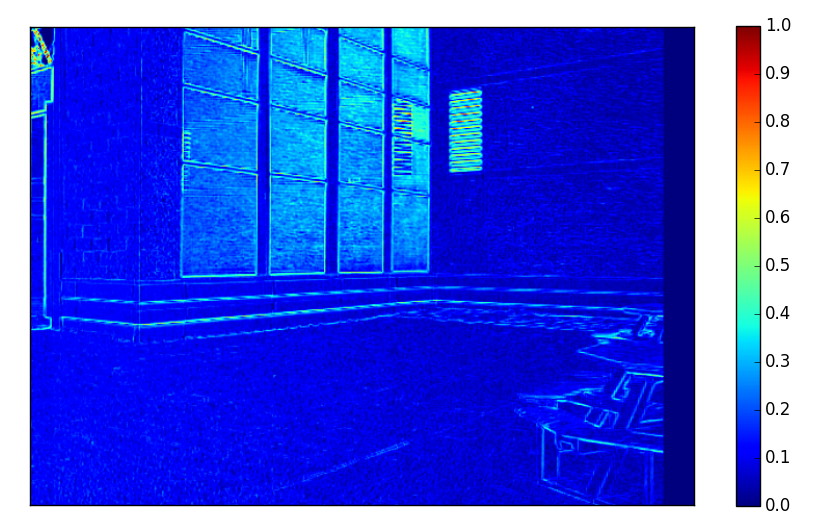}
	\caption{The DOLP of the scene shown in Figure \ref{fig:scene}. The windows have a higher DOLP than the other non specular surfaces in the image.}
	\label{fig:dolp}
\end{figure}

\begin{figure*}[h]
	\centering
	\includegraphics[width=0.9\linewidth]{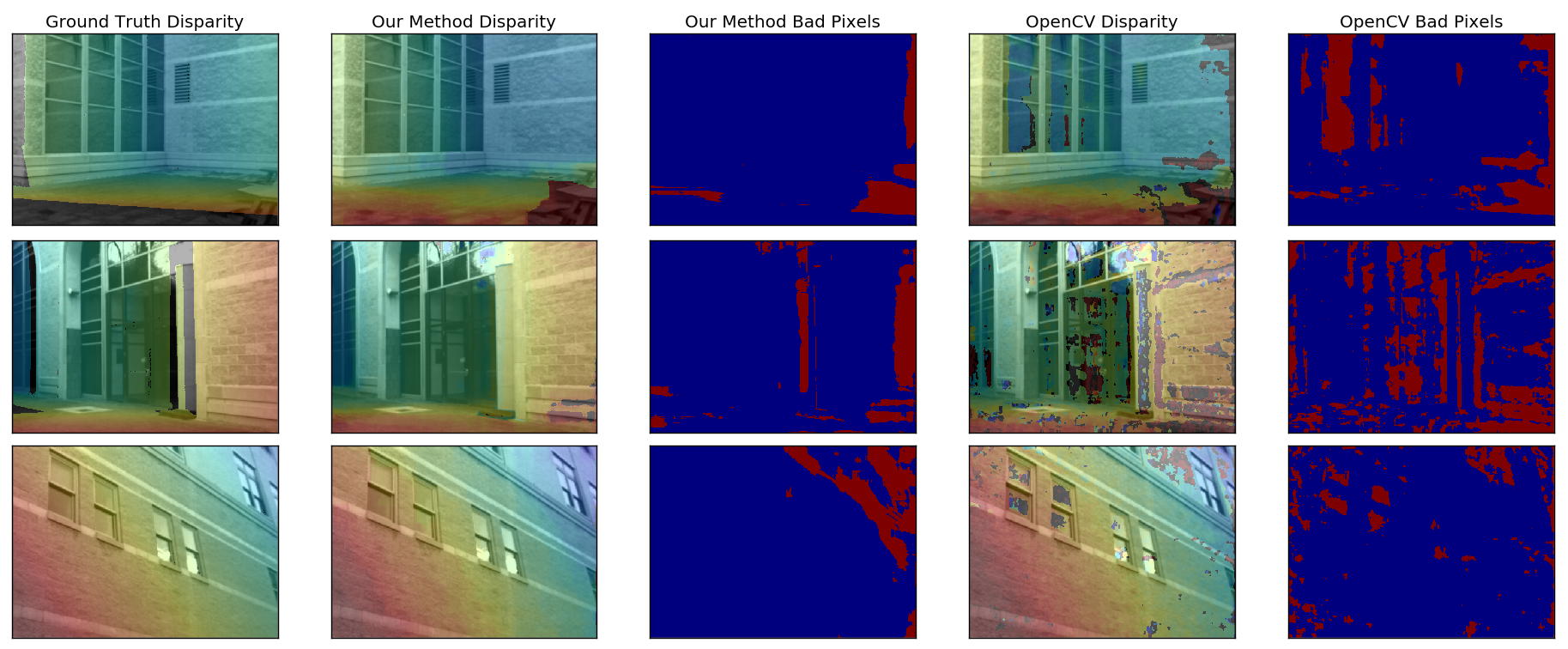}
	\caption{The ground truth disparity, disparity from our method, bad pixels from our method, OpenCV's disparity, and bad pixels from OpenCV on three scenes containing windows.}
	\label{fig:compare}
\end{figure*}

To handle reflective surfaces we use the micropolarizer camera. This is a camera that has a special filter over the image sensor that polarizes light at 0, 45, 90, and 135 degrees. Each group of four adjacent pixels is polarized at one of these polarization angles, which provides enough information to estimate the degree of linear polarization (DOLP) and angle of polarization (AOP). The DOLP indicates highly specular regions. The AOP is used to estimate surface normal in specular regions. An example of the DOLP is shown in Figure \ref{fig:dolp}

We found that the surface normal estimates were too noisy to directly use to interpolate disparity values across high DOLP regions. In \cite{polar}, the noisy surface normal estimates dealt with by incorporating them into their global optimization algorithm. To avoid the computational complexity of a global optimization algorithm, we simply treat continuous regions of high DOLP pixels as planes and fit them into the disparity image by fitting a plane to the adjacent low DOLP subpixel disparity values. Subpixel disparity estimates are obtained by fitting a quadratic curve to the scoring function and finding the minimum value \cite{subpixel}.

\section{RESULTS} \label{results}

We evaluate our approach on data we collected featuring reflective surfaces, objects parallel to the camera baseline, thin obstacles, and repeating patterns.

\subsection{Apparatus}

\begin{figure}[h]
	\centering
    \includegraphics[scale=0.12]{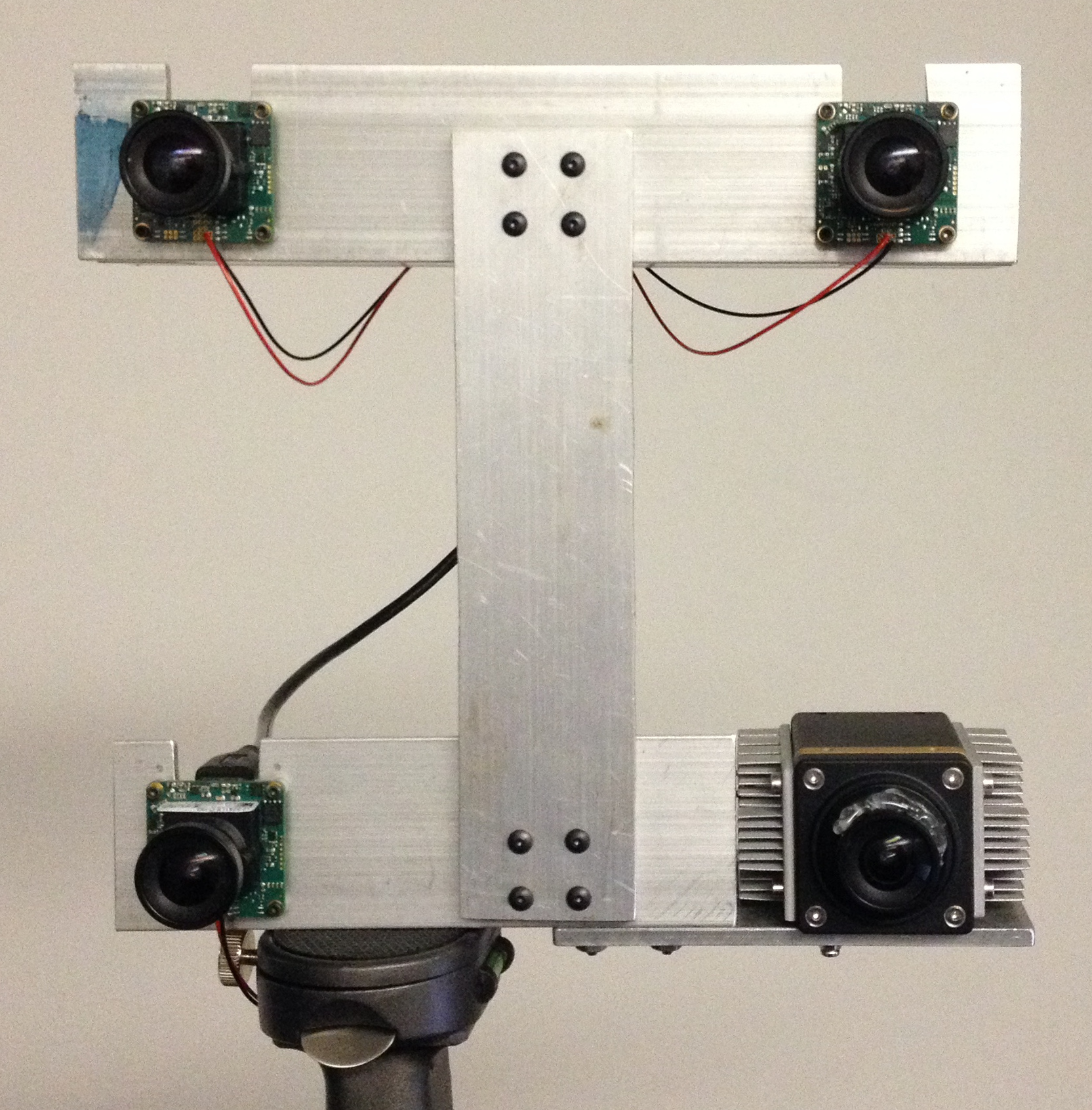}
    \caption{Our trinocular and micropolarizer camera setup. Three uEye UI-3241LE cameras and a 4D Techonology PolarCam V are used.}
    \label{fig:setup}
\end{figure}

Data was collected using the setup shown in Figure \ref{fig:setup}. For trinocular data, three uEye UI-3241LE cameras were used. For stereo data, the 4D Technology PolarCam V and adjacent uEye camera was used. An image resolution of 640x460 was used.

\subsection{Methods}

We compared our stereo algorithm to the commonly used OpenCV's semi global block matching algorithm (SGBM) \cite{opencv}. The same window size was used in both methods. The uniqueness constraint was not applied so that the unrefined output of the disparity algorithms can be compared. Images were collected from three scenes containing windows as shown in Figure \ref{fig:compare}. Ground truth data was collected with a FARO laser scanner. The scanner did not collect accurate data from the windows so we labeled window pixels by hand and estimated their depth by fitting a plane to surrounding non-window measurements. We compared our method against OpenCV using the percentage of bad disparity pixels in the image. Disparity pixels are considered bad if they differ from ground truth by more than two. The percentage of bad pixels in each scene are shown in Table \ref{table:compare}.

\begin{figure*}[h]
	\centering
    \includegraphics[width=0.9\linewidth]{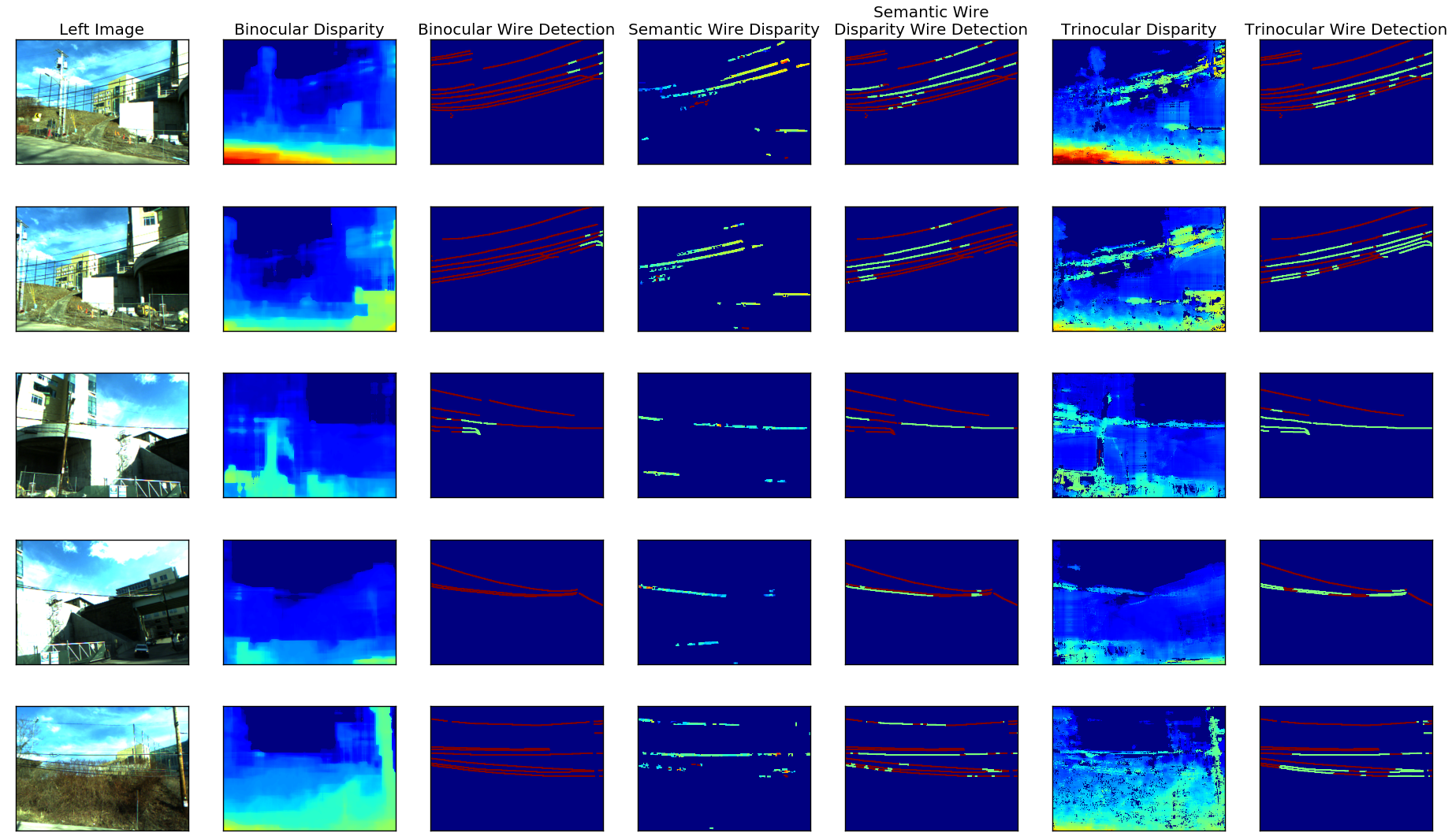}
    \caption{A comparison between binocular disparity, wire detection disparity, and trinocular disparity. The detection results show which wire pixels were detected in green and which were not in red. They have been expanded to be easily visible. Table \ref{table:wires} compares the percentage of wires detected by each method.}
     \label{fig:tri}
\end{figure*}

\subsection{Stereo}

In the first scene, 6.7\% of disparity pixels from our method were bad compared with OpenCV's 17.9\%. Most of the bad pixels in OpenCV came from the windows. Since the windows are reflective, OpenCV is actually estimating the disparity of the wall reflected in the windows, rather than the window themselves. The PolarCam is able to estimate a high DOLP for the window region and our algorithm fits this region into the disparity image. There are also peaks in OpenCV's disparity on the vent next to the windows due to lines parallel to the baseline of the camera. In our method, the additional context from the hierarchical algorithm is able to correctly handle these areas.

In the second scene, 10.8 \% of disparity pixels from our method were bad compared with OpenCV's 27.7\%. Again, most of the bad pixels in OpenCV came from the windows. The top portion of the window reflects trees causing OpenCV to estimate the disparity to be farther than it should. The lower portion of the window is transparent which causes noise in OpenCV's disparity. Our method produces smooth disparity estimates across the window.

In the third scene, 10.3\% of disparity from our method were bad compared with OpenCV's 9.6\%. Our method performed better on the closest windows in the scene. The windows reflect the sky causing them to be textureless which produces noise in OpenCV's disparity. The more distant windows are smaller in the image, so both OpenCV and our method handled them similarly. Our method performed slightly worse in this scene due to the number of bad pixels on the most distant part of the wall. The reason for this is that the depth of the wall at this part of the image is changing more quickly over fewer pixels. Since there is more depth variation in a small area, the hierarchical method incorporates more information from closer depths than farther depths which biases the cost function.

\begin{table}[]
\centering
\begin{tabular}{lllll}
& Scene 1 & Scene 2 & Scene 3  & Mean \\
Our Method & \textbf{6.7} & \textbf{10.8} & 10.3  & \textbf{9.27} \\
OpenCV & 17.9 & 27.7 & \textbf{9.6} & 18.4 \\
\end{tabular}
\caption{Percentage of bad pixels in each scene. A pixel is considered bad if it differs from the ground truth disparity by more than two.}
\label{table:compare}
\vspace*{-10mm}
\end{table}
\subsection{Trinocular}

We also examine the effectiveness our trinocular method in detecting thin obstacles and overcoming inaccuracies due to objects parallel to the cameras' baseline. Figure \ref{fig:tri} shows several scenes featuring wires, some of which are parallel to the horizontal baseline of the cameras. Ground truth data from a laser scanner was not available for these scenes, so instead the wires in the images were labeled by hand. Disparities identified as belonging to the wires were also labeled by hand. Table \ref{table:wires} shows the percentage of wire pixels that each method identified. It also shows the percentage of wire pixels that are detected by combining the semantic and trinocular methods.

\begin{table}[h]
\setlength{\tabcolsep}{2pt}
\centering
\begin{tabular}{lllllll}
& Scene 1 & Scene 2 & Scene 3 & Scene 4 & Scene 5  & Mean\\
Binocular & 3 & 5 & 18 & 0 & 0 & 5\\
Semantic Wire & 36 & 27 & 30 & 26 & 28 & 29\\
Trinocular & 24 & 48 & \textbf{61} & 50 & 35 & 44\\
Merged & \textbf{45} & \textbf{57} & \textbf{61} & \textbf{57} & \textbf{47} & \textbf{53} \\
\end{tabular}
\caption{The percentage of wire pixels detected by the binocular, semantic wire, and trinocular disparity methods. The results from merging the wire detection from the semantic wire and trinocular methods are given in the Merged Disparity row.}
\label{table:wires}
\vspace*{-8mm}
\end{table}

In each case, the binocular camera pair fails to detect the wires, except small portions near the objects they are attached too. The semantic wire method finds disparities for less wire pixels than the trinocular method, however, the sets of wires they detect are different. Since the trinocular method is window based, it inflates the size of the wires and is able to detect nearby wires, leading to higher detection results. In images 1, 2, and 5 the wire disparity algorithm detects the top most wire in the images, which is missed by the weighted trinocular algorithm since it is thinner than the other wires. In scene 4, both methods miss the top most wire, but detect different portions of the thicker wire. Merging the semantic wire and trinocular methods leads to a higher amount of detected wire pixels, with the exception of scene 3,  indicating that the methods are complimentary.

\section{CONCLUSION} \label{conclusion}

In this paper, we presented a method for improving disparity estimation in challenging outdoor environments featuring reflective objects and thin obstacles and overcoming common issues with stereo cameras like repeating patterns and object parallel to the baseline. Our hierarchical method is able to be more robust to repeating patterns, trinocular is able to handle objects parallel to the baseline of one camera pair and thin objects, and reflective regions are identified using the PolarCam and fit into the disparity image.

\section*{Acknowledgements}
This work was supported by Autel Robotics under award number A018532.

\bibliographystyle{unsrt}
\bibliography{references.bib}

\end{document}